**ORIGINAL ARTICLE**

# Learning visual-based deformable object rearrangement with local graph neural networks

Yuhong Deng[1,2] · Xueqian Wang[3] · Lipeng Chen[1]

**Abstract**
Goal-conditioned rearrangement of deformable objects (e.g. straightening a rope and folding a cloth) is one of the most common deformable manipulation tasks, where the robot needs to rearrange a deformable object into a prescribed *goal* configuration with only visual observations. These tasks are typically confronted with two main challenges: the high dimensionality of deformable configuration space and the underlying complexity, nonlinearity and uncertainty inherent in deformable dynamics. To address these challenges, we propose a novel representation strategy that can efficiently model the deformable object states with a set of keypoints and their interactions. We further propose local-graph neural network (GNN), a light local GNN learning to jointly model the deformable rearrangement dynamics and infer the optimal manipulation actions (e.g. pick and place) by constructing and updating two dynamic graphs. Both simulated and real experiments have been conducted to demonstrate that the proposed dynamic graph representation shows superior expressiveness in modeling deformable rearrangement dynamics. Our method reaches much higher success rates on a variety of deformable rearrangement tasks (96.3% on average) than state-of-the-art method in simulation experiments. Besides, our method is much more lighter and has a 60% shorter inference time than state-of-the-art methods. We also demonstrate that our method performs well in the *multi-task* learning scenario and can be transferred to real-world applications with an average success rate of 95% by solely fine tuning a keypoint detector. A supplementary video can be found at https://youtu.be/AhwTQo6fCM0.

**Keywords** Deformable manipulation · Robot learning · Graph neural network · Sim-to-real learning

## Introduction

Deformable objects can be widely seen in many automating tasks such as food handling, assistive dressing and the manufacturing, assembly and sorting of garments [1]. Vision-based deformable object rearrangement (e.g. rope straightening and cloth folding) is one of the most investigated and fundamental deformable manipulation tasks, where the robot is supposed to infer a sequence of manipulation actions (e.g. pick and place) from solely visual observations (e.g. point cloud [2] and RGB image [3]) to rearrange a deformable object into a prescribed goal configuration. Different from rigid manipulation [4–6], deformable rearrangement poses two new challenges. The first challenge lies in the high dimensionality of the deformable configuration space [7]. In contrast to rigid objects, whose configurations are frequently represented as 6-D poses w.r.t. a common reference frame, how to represent the configurations of a deformable object in an efficient and accurate manner, particularly with only visual observations, still remains unresolved. The second challenge comes from the highly complex and non-linear dynamics of deformable materials [8], which makes the object behaviors under certain robot actions (e.g. pull and push) hard to model and predict during manipulation inference and planning.

To tackle high dimensional configuration space, an efficient representation strategy from visual observation of deformable objects is particularly necessary. Convolution Neural Network (CNN) paves the way of extracting hidden

✉ Lipeng Chen
  lipengchen@tencent.com

  Yuhong Deng
  francisdeng@tencent.com

  Xueqian Wang
  wang.xq@sz.tsinghua.edu.cn

1 Tencent Robotics X Lab, Shenzhen, China

2 Present Address: The Center for Intelligent Control and Telescience, Tsinghua Shenzhen International Graduate School, Shenzhen, China

3 The Center for Intelligent Control and Telescience, Tsinghua Shenzhen International Graduate School, Shenzhen, China

dense features of deformable objects from visual observations. Another widely-explored representation strategy is based on keypoint detection from visual observations [9,10]. Representing the states of a deformable object as keypoints can significantly decrease the dimensionality of its configuration space and therefore lead to more data-efficient policy learning of deformable rearrangement while compared with using the convolutional features [11]. However, these methods have the defect of not considering the global interactions among different visual parts, which may convey important clues of deformable configurations. Recently, handcrafted graph structure provide a solution to represent the interactions among keypoints. Concretely, by viewing keypoints as the nodes, the keypoint interactions can be simply represented as edges of a graph structure [12,13]. In our previous work [14], we have used a handcrafted graph structure to represent the deformable object and modified a CNN-based manipulation policy learning architecture. However, handcrafted rules limit the expressiveness of the graph structures in representing deformable configurations. To better handle the complex deformable configurations during robot rearrangement, as an extension of our previous work [15], we proposed a learned dynamic graph representation strategy where the interactions among keypoints are learned rather than predefined with handcrafted rules. The keypoint interactions are learned by the model in the rearranging policy learning. Since the graph is constructed to improve the performance of rearrangement tasks, the learned keypoint interactions go beyond the expert knowledge and therefore can be more suitable and efficient for representing the deformable object configurations.

As for manipulation policy learning under deformable dynamics, the end-to-end fashion have become a research focus recently, where the robot learns deformable rearrangement policy from visual observation recently. However, most pioneering works tend to provide task-specific models. Research on establishing a general framework that can be used on different rearranging tasks has achieved some progress recently. Seita et al. [3] proposed the goal-conditioned transporter network which performs well on several deformable rearranging tasks. Lim et al. [16] present a more systematic classification of the rearrangement tasks and achieved a better performance in multi-task learning. However, the end-to-end policy learning from visual observation depends heavily on the image style and therefore can potentially bring about severe sim-to-real learning gaps. To narrow the sim-to-real gap, we propose a two-staged learning framework (Fig. 1). More specifically, we firstly design a keypoint detector that can extract keypoints from visual observations effectively. And we propose local-GNN, a local graph neural network to learn manipulation policy from keypoints in current and goal visual observations. In this way, the image style is isolated from manipulation policy learning. The current

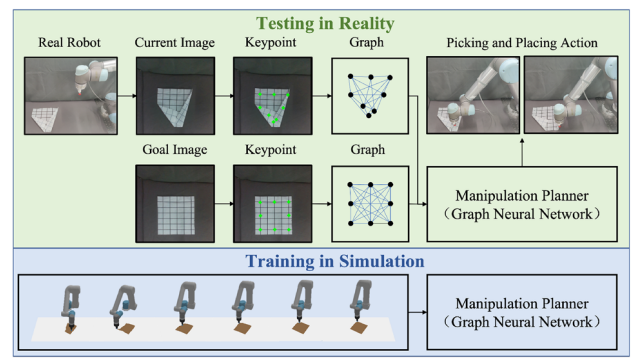

**Fig. 1** Illustration of the proposed local-CNN. We propose local-GNN, a light and suitable local graph neural network learning to manipulate deformable objects with an efficient and accurate graph representation of the deformable rearrangement dynamics

and goal keypoints are transformed into two dynamic graph, and proposed local-GNN first updates both graphs locally to obtain the accurate representations of object states during robot rearrangement. Then proposed local-GNN exchanges messages globally across the updated local graphs to find their best-matched node pairs, i.e. to search for the optimal manipulation action pairs that can narrow the gap between the current and goal state to the best.

We present both simulated and real experiments on a variety of deformable rearrangement tasks to evaluate our proposed method. The results demonstrate that our method can be a more efficient and general for vision-based goal-conditioned deformable rearrangement compared with state-of-the-art frameworks. Moreover, leveraging the keypoint and graph representation, our model is more expressive in deformable rearrangement dynamics, but much lighter in model size and complexity. Besides *single-task* learning, our method achieves comparable *multi-task* learning performance. Real-world experiments also reveal the enhanced sim-to-real transferability of our framework. The contributions of this work are summarized as follows:

(1) We propose a novel graph representation strategy, where the keypoints are detected and encoded into a dynamic graph as an efficient and accurate structural replacement of the high-dimensional visual observations for vision-based deformable rearrangement;
(2) We propose local-GNN, a light and effective graph neural network that utilizes local and global attentions between two dynamic graphs to automatically learn the deformable rearrangement dynamics and to infer the optimal deformable rearranging policy together;
(3) We propose the two-staged learning framework, which greatly narrows the sim-to-real gap of learning vision-based goal-conditioned deformable rearrangement.

The rest of this paper is organized as follows. The related work is reviewed in "Related work". We introduce our learn-ing framework briefly in "Problem formulation". "Learning for deformable object rearrangement" presents details of learning algorithms in our proposed framework. The experimental setup and results are provided/in "Experiment results". "Conclusion" concludes this paper and discusses some future works.

## Related work

This section reviews related work on the configuration representation of deformable objects, manipulation planning of deformable rearrangement, and manipulation policy learning from keypoints (Table 1).

### Configuration representation of deformable objects

Considering the high dimensionality of the configuration space of deformable objects, an effective representation method is necessary. Adopting Convolution Neural Network (CNN) to extracting hidden dense features of deformable objects from visual observations is widely investigated. However, CNN focuses solely on the local features in visual images, which limits its expressiveness and efficiency in deformable object configurations, as the global relationships (or interactions) among different parts of a deformable object is important. Besides, the convolutional features depend heavily on the image style and therefore can potentially bring about severe sim-to-real learning gaps. Keypoints representation strategies have the advantage of lower dimensionality of configuration space compared with CNN feature. Considering that keypoint representation methods also ignore interactions between keypoints, Miller et al. [22] introduced predefined geometric constraints to incorporate keypoint interactions, which however are a strong prior and can hardly be obtained accurately. Recent advances in graphs provide another potential solution to represent the interactions among keypoints without geometric prior. However, keypoint interactions in these graphs are usually handcrafted, i.e. each interaction is simply constructed as whether there exists a physical connection between two keypoints and each keypoint is connected to its neighbors within a predefined distance. In our model design, keypoint interactions is learning during rearrangement policy learning instead of defining by handcrafted rules, which can exploit the potential of graphs in rearrangement more fully.

### Manipulation planning of deformable objects

There are two main approaches towards manipulation planning of deformable objects. Particularly, model-based approaches rely on an accurate forward dynamic model, which can predict the configurations of deformable objects under a certain manipulation actions. The forward dynamic modeling method can be mainly divided into the physics-based methods (e.g. mass-spring system and continuum mechanics) and the data-driven methods. The accuracy of physics-based methods highly depends on tuning the involved physical parameters properly and therefore an accurate physics-based method is usually too complex and expensive to obtain [1]. The data-driven methods learn the forward dynamic model from quantities of interaction data between robots and deformable objects [19], which however is data-consuming and poor in generalization on different objects and tasks.

Policy-based approaches aim to obtain optimal manipulation policies directly from observation without the establishment of a forward dynamic model. This line of works can be divided into two categories according to the source of the training data: imitation learning and reinforcement learning. In imitation learning, the manipulation task is often formulated as a supervised learning problem where the robot should imitate the observed behaviors [18]. Reinforcement learning obtains rearranging skills through robot exploratory interactions [17,23]. However, most previous policy-based methods are limited to *single-task* learning, which is inefficient in real-world applications. However, our model learns manipulation policies from dynamic graphs with an efficient local-GNN architecture, which our proposed method is proved be a general framework for multiple deformable object rearrangement tasks and performs well in the *multi-task* learning scenario.

### Manipulation policy learning from keypoints

Learning manipulation policies from keypoints has become a research focus recently because keypoints can be effective alternatives for high-dimensional visual observations and lead to more data-efficient manipulation policy learning. Within the context of deformable object manipulation, early works focus on tracking the keypoints of deformable objects [24]. To achieve efficient manipulation policy learning, Lin et al. [25] have used the positions of keypoints on the rope as the reduced states. To bridge the sim-to-real gap, Wang et al. [20] treated keypoints as nodes in a graph and designed an offline-online learning framework based on graph neural networks. Ma et al. [21] designed a graph neural network to learn the forward dynamic model of the deformable objects and achieved precise visual manipulation. However, most previous graph neural network-based methods rely on a model predictive controller to compensate for the prediction error of the forward dynamic model, which brings a heavy computational burden. In addition, the generalization of the pre-trained dynamic model on different objects and tasks is usually limited. To provide a general learning framework, we design a local-GNN to learn manipu-lation policy directly from keypoints (represented as a graph) without the establishment of a forward dynamic model.

**Table 1** Research gaps and contributions of previous studies

| Author(s) | Method type | Object | Representation strategy | Sim2real strategy | Policy learning strategy |
| --- | --- | --- | --- | --- | --- |
| Matas et al. [17] | Policy based | Cloth | Convolutional visual features | Domain randomization in simulation | Single-task learning |
| Nair et al. [18] | Policy based | Rope | Convolutional visual features | Collecting data on the real robot | Single-task learning |
| Seita et al. [3] | Policy based | Rope, cloth and bag | Convolutional visual features | NA | Single-task learning |
| Lim et al. [16] | Policy based | Rope | Convolutional visual features | NA | Multi-task learning |
| Yan et al. [19] | Model-based | Rope, cloth | Latent space vector | Collecting data on the real robot | Single-task learning |
| Wang et al. [20] | Model-based | Rope | Graph feature | Residual model | Single-task learning |
| Ma et al. [21] | Model-based | Rope, cloth | Graph feature | Collecting data on the real robot | Single-task learning |
| Our method | Policy-based | Rope, cloth | Graph feature | Transfer to the real robot directly | Multi-task learning |

## Problem formulation

This section presents a detailed problem definition and a brief introduction of the proposed learning framework. Table 2 lists key notations used in this work.

### Problem formulation

As shown in Fig. 2, given a visual observation $\mathbf{I}_0 \in \mathbb{R}^{w \times h \times c}$ of the initial state of the deformable object and a prescribed goal state $\mathbf{I}_g \in \mathbb{R}^{w \times h \times c}$ also specified as a visual observation, where w, h and c denote the width, height and channel of the visual observations respectively, we formulate the problem of goal-conditioned deformable rearrangement as to find a policy $\pi$ that can generate a sequence of robot *pick* and *place* actions $\{\alpha_t\}(t = 0, 1, 2, \ldots k)$ in a closed-loop manner:

$$\alpha_t \leftarrow \pi(\mathbf{I}_t, \mathbf{I}_g) \text{ and } \mathbf{I}_{t+1} \leftarrow \mathcal{T}(\mathbf{I}_t, \alpha_t) \quad (1)$$

such that:

$$\|\mathbf{I}_{k+1} - \mathbf{I}_g\|_{\text{latent}} \leq \gamma \quad (2)$$

where $\mathcal{T}$ denotes the state transition describing the deformable rearrangement dynamics to be learned by the policy, and $\gamma$ denotes the similarity threshold defined in the latent space, which determines if the object state is close enough to the goal state during robot manipulation.

We define each action $\alpha \in \mathcal{A}$ as a *pick* action followed by a *place* action:

$$\alpha_t = \{\alpha_t^{\text{pick}}, \quad \alpha_t^{\text{place}}\} \quad (3)$$

where $\alpha_t^{\text{pick}}$ and $\alpha_t^{\text{place}}$ denote the poses of robot end-effector while grasping and releasing part of the object respectively.

**Table 2** Nomenclature

| Symbol | Description |
| --- | --- |
| **I** | The visual observations of a deformable object |
| $\boldsymbol{\alpha}$ | The robot pick and place action |
| $\mathcal{P}$ | A set of keypoints detected from the visual observation |
| $p$ | A keypoint detected from the visual observation |
| **H** | A feature map extracted from the visual observation |
| $\Omega$ | The pixel domain of the visual observation |
| $\Omega'$ | The pixel domain of the feature map |
| $\mathbf{G}(\mathcal{P})$ | Gaussian heatmaps centred at the keypoints $\mathcal{P}$ |
| $\mathbf{Q}(\mathcal{P})$ | The probability distribution of action success on the keypoints $\mathcal{P}$ |

More concretely, we consider tabletop manipulation tasks and hence both poses $\alpha_t^{\text{pick}}$ and $\alpha_t^{\text{place}}$ are defined in SE(2), where positions are sampled from a fixed-height planar surface ($x$–$y$) and rotations are defined around the $z$-axis. Similar to previous work [3,26], our method obtains the pick or place height ($z$) w.r.t. a heightmap generated from visual observations.

### Method overview

Our policy $\pi$ consists of two main components, namely a keypoint detector $\pi_{\text{keypoint}}$ that extracts two sets of keypoints, denoted by $\mathcal{P}_t, \mathcal{P}_g$, from the current and goal visual observations $\mathbf{I}_t, \mathbf{I}_g$ of a deformable object, respectively, and a manipulation planner $\pi_{\text{plan}}$ that determines optimal pick and place actions on the keypoints from sets $\mathcal{P}_t, \mathcal{P}_g$, to manipulate the object to the goal state $\mathbf{I}_g$ in a close loop manner.

Concretely, the keypoint detector $\pi_{\text{keypoint}}$ firstly maps from the high-dimensional visual observations $\mathbf{I}_t, \mathbf{I}_g$ of the deformable object into two sets of 2-D keypoints $\mathcal{P}_t, \mathcal{P}_g$,

$$\mathcal{P}_t \leftarrow \pi_{\text{keypoint}}(\mathbf{I}_t), \mathcal{P}_g \leftarrow \pi_{\text{keypoint}}(\mathbf{I}_g) \quad (4)$$

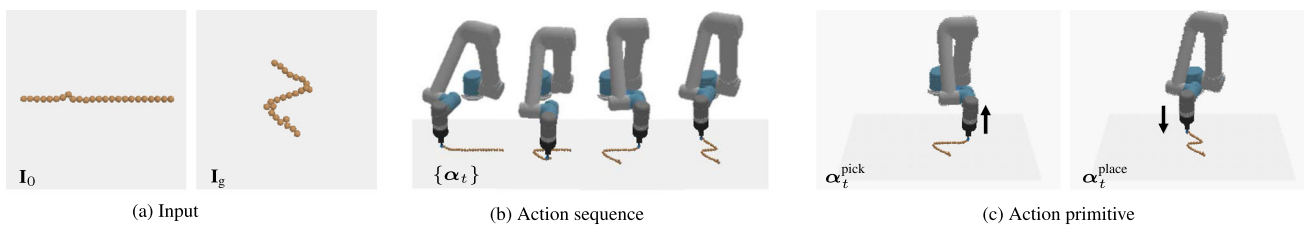

(a) Input  (b) Action sequence  (c) Action primitive

**Fig. 2** Problem overview. We formulate the goal-conditioned deformable rearrangement problem as to find a sequence of robot pick and place actions to rearrange the object from an initial state $\mathbf{I}_0$ to a prescribed goal state $\mathbf{I}_g$. **a** The input contains only the visual observations $\mathbf{I}_0$, $\mathbf{I}_g$ of the current and goal object states. **b** The method generates a sequence of pick and place actions $\{\boldsymbol{\alpha}_t\}$ ($t = 0, 1, 2, \ldots k$) rearranging the deformable object to the goal state. **c** Each action is defined as a pair of a pick $\boldsymbol{\alpha}_t^{\text{pick}}$ and a place $\boldsymbol{\alpha}_t^{\text{place}}$ action

where each keypoint $\boldsymbol{p}$ in $\mathcal{P}_t$, $\mathcal{P}_g$ locates a candidate position for robot *pick* and *place* on the object respectively. We aim to train the keypoint detector such that the output keypoints $\mathcal{P}$ from the visual observation $\mathbf{I}$ can be a more efficient, accurate and compact structural representation of the deformable object facilitating the subsequent searching in planning ("Learning for keypoint detection").

The planner $\pi_{\text{plan}}$ then reasons from the keypoints to find the optimal pair of pick and place actions

$$\boldsymbol{\alpha}_t = \pi_{\text{plan}}(\mathcal{P}_t, \mathcal{P}_g). \tag{5}$$

We formulate the manipulation planning of deformable rearrangement as a sequence-to-sequence (S2S) problem, and propose the local-GNN to learn the deformable rearrangement dynamics and reason about the optimal pick and place actions until Eq. (2) is satisfied ("Learning for deformable rearrangement").

## Learning for deformable object rearrangement

This section presents details of our learning framework for goal-conditioned deformable object rearrangement. Briefly, it consists of extracting keypoint and graph representations and then finding the optimal manipulation actions from visual observations.

### Learning for keypoint detection

Our method starts from extracting a set of keypoints $\mathcal{P} = \{\boldsymbol{p}_i\}_{i=1}^{m}$ from the visual observation $\mathbf{I}$, as an effective accurate and compact structural representation of the high-dimensional configuration of the deformable object. We borrow ideas from previous work [27,28] and train a deep convolution neural network as our keypoint detector $\pi_{\text{keypoint}}$. As shown in Fig. 3, given a visual observation $\mathbf{I} \in \mathbb{R}^{\text{h} \times \text{w} \times \text{c}}$ of the deformable object, the detector first produces a group of $m$ feature maps $\{\mathbf{H}_i\}_{i=1}^{m}$ from the visual cues extracted in the latent space. Each feature map $\mathbf{H}_i \in \mathbb{R}^{\text{h}' \times \text{w}'}$ is then regarded as a map of probability distribution processed to find the exact location $\boldsymbol{p}_i$ of the $i$-th corresponding keypoint in the visual observation.

We apply a spatial softmax strategy to extract the keypoint location from a feature map. Specifically, given a feature map $\mathbf{H}_i$, while using $\Omega : \text{w} \times \text{h}$ and $\Omega' : \text{w}' \times \text{h}'$ to denote the pixel domain of the visual observation $\mathbf{I}$ and the feature map $\mathbf{H}_i$ respectively, the location of its corresponding $i$-th keypoint on the feature map domain $\Omega'$ can be obtained as

$$\boldsymbol{p}_i^{\Omega'} = (x_i^{\Omega'}, y_i^{\Omega'}) = \frac{\sum_{\boldsymbol{c} \in \Omega'} \boldsymbol{c} \times \exp(\mathbf{H}_i(\boldsymbol{c}))}{\sum_{\boldsymbol{c} \in \Omega'} \exp(\mathbf{H}_i(\boldsymbol{c}))} \tag{6}$$

where $\boldsymbol{c} \in \Omega'$ denotes a pixel location on the feature map. The superscript indicates the reference domain and is omitted for the visual observation domain $\Omega$ for simplicity. The spatial softmax strategy (Eq. (6)) condenses each feature map into a keypoint, which is fully differentiable and therefore makes the keypoint detector trainable.

The corresponding keypoint $\boldsymbol{p}_k$ on the visual observation domain $\Omega$ can then be obtained via a linear scaling mapping from the keypoint on feature map domain $\boldsymbol{p}_i^{\Omega'}$

$$\boldsymbol{p}_k = (x_k, y_k) = \left(\frac{\text{H}}{\text{H}'} x_k^{\Omega'}, \frac{\text{W}}{\text{W}'} y_k^{\Omega'}\right). \tag{7}$$

We train the keypoint detector $\pi_{\text{keypoint}}$ by optimizing over a Gaussian heatmap in a supervised fashion. Specifically, rather than optimizing the detector directly on the keypoints, our method generates two Gaussian heatmaps centered at the estimated keypoints $\mathcal{P} = \{\boldsymbol{p}_i\}_{i=1}^{m}$ and at their corresponding ground truth locations $\mathcal{P}^* = \{\boldsymbol{p}_i^*\}_{i=1}^{m}$ on the visual observation respectively

$$\mathbf{G}(\mathcal{P}) = \sum_{i=1}^{m} \exp\left(-\frac{1}{2\sigma^2} \|\boldsymbol{p} - \boldsymbol{p}_i\|^2\right), \quad \boldsymbol{p} \in \Omega \tag{8}$$

$$\mathbf{G}^*(\mathcal{P}^*) = \sum_{i=1}^{m} \exp\left(-\frac{1}{2\sigma^2} \|\boldsymbol{p} - \boldsymbol{p}_i^*\|^2\right), \quad \boldsymbol{p} \in \Omega \tag{9}$$

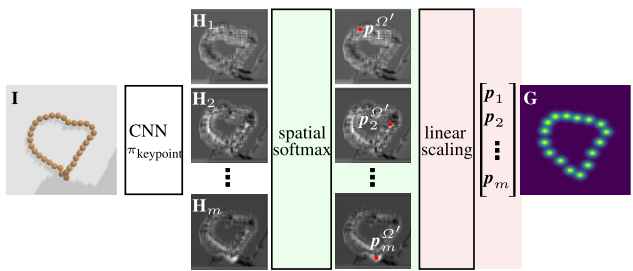

**Fig. 3** Keypoint detector. We design a keypoint detector to exact keypoints from the RGB observations. The keypoint detector outputs a Gaussian heatmap centered at the coordinates of keypoints from the RGB observations

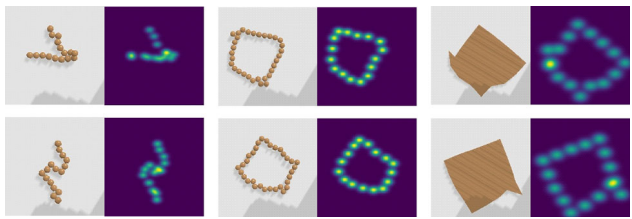

**Fig. 4** Gaussian heatmaps generated by our keypoint detector. The left column represents the original visual observation, and the right column represents the Gaussian heatmap centered at keypoints detected from the visual observation

where $\sigma$ is the constant standard deviations. The keypoint detector is then trained by minimizing a pixel-wise L2 loss between the Gaussian heatmaps $\mathbf{G}$ and $\mathbf{G}^*$. The Gaussian representation of keypoints provides additional information on pixels which are less likely to be keypoints, and therefore make the training more efficiently.

Figure 4 shows example Gaussian heatmaps generated by our keypoint detector. Leveraging the keypoint detector, our method transforms each visual observation into a set of keypoints, which are then fed into the manipulation planner as candidate locations for robot pick and place. Rather than searching in the whole pixel domain of visual observations, where most locations are either invalid or redundant for deformable rearrangement, our method reduces the action exploration into a limited number of keypoints, making the subsequent planning more efficient.

## Learning for deformable rearrangement

Our method then reasons from the keypoints of the current and goal states $\mathcal{P}_t, \mathcal{P}_g$ at each time step to find the optimal pick and place actions respectively, i.e. $\alpha_t^{\text{pick}} \in \mathcal{P}_t$, $\alpha_t^{\text{place}} \in \mathcal{P}_g$ ($t = 0, 1, 2, \ldots, k$), which is formulated as a sequence-to-sequence problem

$$\mathbf{Q}_{\text{pick}}, \mathbf{Q}_{\text{place}} = \psi(\mathcal{P}_t, \mathcal{P}_g) \tag{10}$$

where $\mathbf{Q}_{\text{pick}}, \mathbf{Q}_{\text{place}}$ correspond to the probability distributions of pick and place success on the keypoints in $\mathcal{P}_t$ and $\mathcal{P}_g$ respectively. The optimal pick and place actions can therefore be determined as

$$\begin{aligned}
\alpha_t^{\text{pick}} &= \underset{p \in \mathcal{P}_t}{\arg\max}(\mathbf{Q}_{\text{pick}}(p)) \\
\alpha_t^{\text{place}} &= \underset{p \in \mathcal{P}_g}{\arg\max}(\mathbf{Q}_{\text{place}}(p)).
\end{aligned} \tag{11}$$

We build local-GNN with an *attention*-based updating strategy to learn the above functions (Eqs. (10) and (11)). Briefly, leveraging the two sets of keypoints, our method first constructs two dynamic graphs to represent the current and goal states of the deformable object. The two graphs are then further updated by local-GNN (1) to exact the hidden features that can effectively characterize the object states, (2) to learn the deformable rearrangement dynamics, and (3) to infer the optimal pick and place actions that can drive the object from the current state to the goal state as close as possible, from the intrinsic structures and underlying interactions among the keypoints/nodes of the two dynamic graphs.

**Model architecture:** Concretely, as shown in Fig. 5, at each time step, the keypoints $\{\mathcal{P}_t, \mathcal{P}_g\}$ detected from the current and goal visual observations $\{\mathbf{I}_t, \mathbf{I}_g\}$ are first embedded into a latent space with a multilayer perceptron (MLP),

$$x^0 = \text{MLP}^0(p). \tag{12}$$

The obtained hidden features $\{\mathcal{X}_t^0, \mathcal{X}_g^0\}$ are then utilized as the initial nodes of two dynamic graphs $\{\mathcal{G}_t, \mathcal{G}_g\}$ to represent the initial and goal states of the deformable object respectively. Particularly, we define two types of graph edges (Fig. 5-Right), including *self-edges* $\epsilon_s$ connecting every two nodes from a same graph and *cross-edges* $\epsilon_c$ connecting every two nodes from across two different graphs.

The two graphs are then passed and updated progressively through local-GNN, which consists of a number of $\ell_s$ self-attention layers for local graph updating, and a number of $\ell_c$ cross attention layers for keypoint matching between graphs. Specifically, at each updating layer, all nodes in $\{\mathcal{X}_t, \mathcal{X}_g\}$ are updated by aggregating messages through self-edges $\mathcal{E}_s$ or cross-edges $\mathcal{E}_c$ in the two graphs respectively:

$$x_i^{\ell+1} = x_i^\ell + \text{MLP}\left(\left[x_i^\ell \| m_{\mathcal{E} \to i}\right]\right) \tag{13}$$

where the right superscript indexes the updating layer, and $[\cdot\|\cdot]$ denotes the concatenation operation. The updating message $m_{\mathcal{E} \to i}$ for the $i$-th node represents the resultant messages aggregated from nodes $\{j : \epsilon_{j \to i} \in \mathcal{E}\}$ connected to the $i$-th node, where the edge set $\mathcal{E} = \mathcal{E}_s$ and $\mathcal{E} = \mathcal{E}_c$ during the local and cross updating stage respectively.

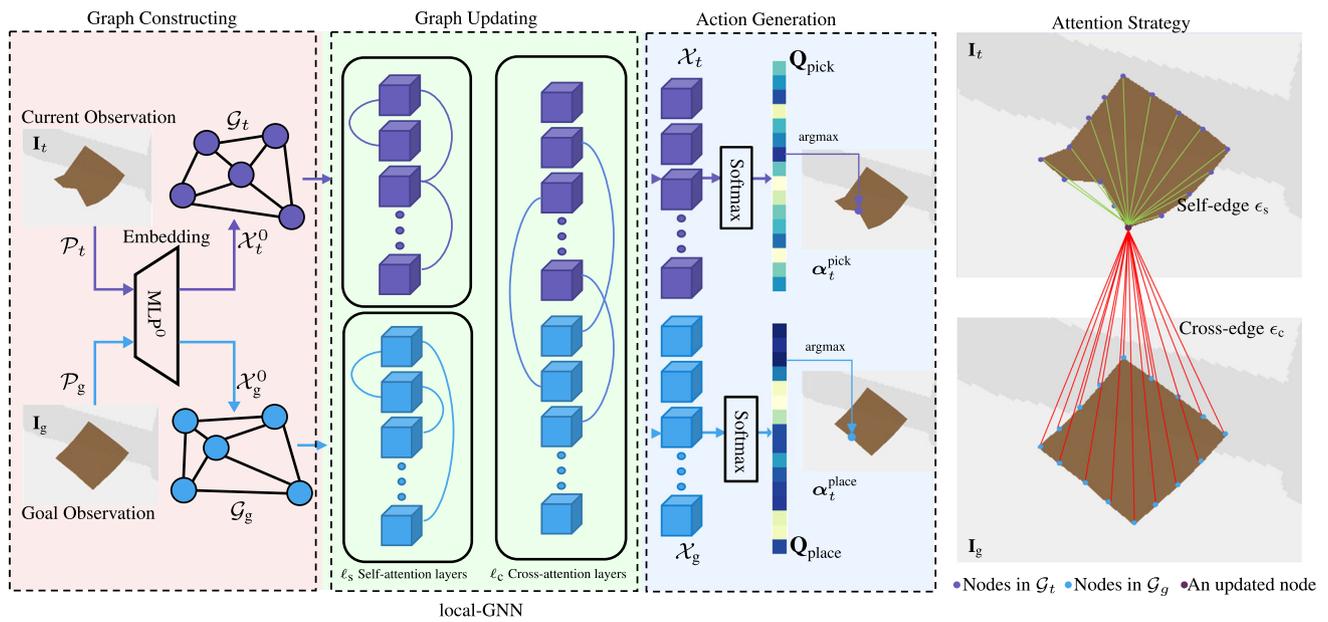

**Fig. 5** Model architecture. Our model detects keypoints from current and goal visual observations and establish the representation vectors of keypoints firstly. Two representation graphs are established based on keypoints. A local-GNN with self-attention layers and cross-attention layers is used to update the two graphs then. Finally, the probability distribution of *pick* and *place* action is generated from the updated graph nodes by a soft-max layer

Our method aggregates the message $m_{\mathcal{E} \to i}$ for the $i$-th node by using its attention values with all neighboring nodes as in Transformer [29], which has been widely applied for sequence-to-sequence processing [30,31]. Specifically, at each self-attention layer ($\ell \leq \ell_s$, $\mathcal{E} = \mathcal{E}_s$), the network aggregates messages through self-edges to obtain a more accurate and efficient graph representation of the object state in rearrangement. While at each cross-attention layer ($\ell_s < \ell \leq \ell_c$, $\mathcal{E} = \mathcal{E}_c$), the network aggregates messages through cross-edges to further update both graphs $\mathcal{G}_t$ and $\mathcal{G}_g$, which essentially compares and matches nodes between $\mathcal{X}_t$ and $\mathcal{X}_g$ to search for the optimal pair of pick and place actions.

Finally, the updated graph nodes are passed through a soft-max layer to output the probability distributions of pick and place actions $\mathbf{Q}_{\text{pick}}$, $\mathbf{Q}_{\text{place}}$ on their corresponding keypoints $\mathcal{P}_t$, $\mathcal{P}_g$ respectively, from which the robot selects the optimal pick and place actions to rearrange the deformable object as given by Eq. (11). The above procedure runs in a close loop manner until the obtained object state is close enough to the prescribed goal state.

### Training and implementation

To train the network, we adopt the paradigm of imitation learning and generate a dataset of $N$ stochastic expert demonstrations $\mathcal{D} = \{\tau_i\}_{i=1}^N$, where each demonstration $\tau_i$ contains a sequence of visual observation and action pairs:

$$\tau_i = \{(\mathbf{I}_1, \boldsymbol{\alpha}_1), (\mathbf{I}_2, \boldsymbol{\alpha}_2), \ldots, (\mathbf{I}_{T_i}, \boldsymbol{\alpha}_{T_i})\}. \tag{14}$$

Details of the demonstration dataset are present in "Dataset construction". Leveraging the dataset, we formulate the training of local-GNN as a supervised classification problem. We employ the cross-entropy error as the loss function, which has been widely proved to be efficient for classification learning:

$$\mathcal{L} = w_{\text{pick}} * \mathcal{L}_{\text{pick}} + w_{\text{place}} * \mathcal{L}_{\text{place}} \tag{15}$$

and

$$\mathcal{L}_{\text{pick}} = -\sum_{\boldsymbol{p} \in \mathcal{P}_{\text{pick}}} y_{\text{pick}}(\boldsymbol{p}) \log(\mathbf{Q}_{\text{pick}}(\boldsymbol{p}))$$

$$\mathcal{L}_{\text{place}} = -\sum_{\boldsymbol{p} \in \mathcal{P}_{\text{place}}} y_{\text{place}}(\boldsymbol{p}) \log(\mathbf{Q}_{\text{place}}(\boldsymbol{p})) \tag{16}$$

where $\mathcal{L}_{\text{pick}}$, $\mathcal{L}_{\text{place}}$ denote the pick and place loss, and $w_{\text{pick}}$, $w_{\text{place}}$ are the scalar coefficients of loss weight respectively. If $\boldsymbol{p}$ is a ground truth pick/place position, $y_{\text{pick/place}}(\boldsymbol{p}) = 1$, otherwise $y_{\text{pick/place}}(\boldsymbol{p}) = 0$.

### Dataset construction

Learning an end-to-end robot manipulation policy is particularly data-thirsty. However, large-scale datasets of real-robot demonstrations are usually costly and inaccessible. To this end, we modify the simulation environment in [3] and construct our own dataset of deformable rearrangement tasks in PyBullet [32], which provides highly-realistic visual rendering and simulations of deformable dynamics.

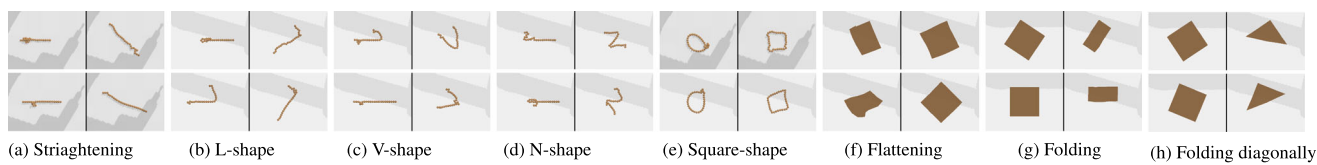

(a) Straightening (b) L-shape (c) V-shape (d) N-shape (e) Square-shape (f) Flattening (g) Folding (h) Folding diagonally

**Fig. 6** Deformable rearrangement tasks in our dataset. Each represents two instances, where the left are the current states and the right are the goal states

**Table 3** Deformable rearrangement tasks involved in our dataset

| Dimension | Categories | Num | Tasks |
|---|---|---|---|
| 1D | Rope | 1100 | Straightening (Fig. 6a) |
| 1D | Rope | 1100 | L-shape (Fig. 6b) |
| 1D | Rope | 1100 | V-shape (Fig. 6c) |
| 1D | Rope | 1100 | N-shape (Fig. 6d) |
| 1D | Rope ring | 1100 | Square shape (Fig. 6e) |
| 2D | Cloth | 1100 | Flattening (Fig. 6f) |
| 2D | Cloth | 1100 | Folding (Fig. 6g) |
| 2D | Cloth | 1100 | Folding diagonally (Fig. 6h) |

Particularly, we set up a fixed UR5 manipulator mounted with a suction cup as its end-effector for deformable rearrangement. We use a RGB-D camera just above the robot workspace to collect visual observations of deformable objects from a top-down perspective. We build the dataset with three types of deformable objects, including ropes, rope rings and cloth (Fig. 6). Each type of deformable objects is designed with a large number of rearrangement tasks with great randomness, such that the task diversity in our dataset is sufficient to guarantee the generalization and effectiveness of the learned framework. It is worth noting that both initial and goal configurations of our rearrangement tasks are completely randomized and provided as visual observations, which poses a higher generalization requirement to our framework. Details of the deformable rearrangement tasks in our dataset are present in Table 3 and in the supplementary materials.

## Experiment results

This section presents both simulated and real-world experiments to evaluate the performance of our framework. Particularly, we aim to answer the following questions: (1) How well does our framework compare with the baseline methods on deformable rearrangement tasks? (2) How well does our framework perform on real-world deformable rearrangement tasks? and (3) How well does our framework perform on *multi-task* policy learning?

### Simulation experiments

We first evaluate the performance of our method on a variety of goal-conditioned deformable rearrangement tasks in simulation. Specifically, at each experiment of rearranging a certain deformable object, the robot is provided with a random goal state of the object in the form of visual image, and is supposed to manipulate the object to the goal state without any intermediate sub-goal inputs.

### Baseline methods

We compared our method with five baseline methods widely applied to goal-conditioned deformable rearrangement:

1. Conv-MLP represents a convolution-based neural network model architecture, which consists of a group of convolutional layers followed by a multilayer perceptron (MLP) to regress the robot pick and place actions from visual observations directly.
2. GTCN [3] represents the typical goal-conditioned transporter network for deformable rearrangement. It relies not only on the convolutional operations to exact hidden dense features from visual observations, but also leverages the cross-correlation between dense convolution features (namely the transporter operation) of the current and goal states to infer optimal pick and place actions.
3. Graph Transporter [14] represents an optimized GCTN network, where handcrafted graph structure is used to represent deformable objects and supplement global interaction information to CNN features during manipulation policy learning.
4. MLP represents a multilayer perceptron, which learns manipulation policy directly from positions of keypoints.
5. GCN represents a Graph Convolutional Network architecture, which learns manipulation policy directly from the positions of keypoints. In GCN, each node is updated by aggregating messages from edges in the graph. The (edges) node connection relationships are handcrafted in GCN. We define a complete graph, where each node is connected to all other nodes.

**Table 4** Results of success rates

| Method | Straightening | | | L-shape | | | V-shape | | | N-shape | | |
|---|---|---|---|---|---|---|---|---|---|---|---|---|
| | 10 | 100 | 1000 | 10 | 100 | 1000 | 10 | 100 | 1000 | 10 | 100 | 1000 |
| Conv-MLP | 0.0 | 5.0 | 12.5 | 0.0 | 2.5 | 5.0 | 0.0 | 2.5 | 7.5 | 0.0 | 5.0 | 7.5 |
| GCTN | 2.5 | 20.0 | 75.0 | 0.0 | 5.0 | 60.0 | 0.0 | 10.0 | 62.5 | 0.0 | 10.0 | 40.0 |
| Graph transporter | 37.5 | 70.0 | 75.0 | 15.0 | 35.0 | 45.0 | 5.0 | 10.0 | 65.0 | 5.0 | 15.0 | 50.0 |
| MLP | 0.0 | 0.0 | 5.0 | 2.5 | 5.0 | 7.5 | 0.0 | 5.0 | 12.5 | 0.0 | 5.0 | 5.0 |
| GCN | 0.0 | 10.0 | 15.0 | 0.0 | 10.0 | 10.0 | 2.5 | 7.5 | 15.0 | 0.0 | 0.0 | 5.0 |
| Local-GNN (ours) | **45.0** | **95.0** | **100.0** | **60.0** | **70.0** | **100.0** | **55.0** | **65.0** | **100.0** | **20.0** | **45.0** | **85.0** |
| Method | Square shape | | | Flattening | | | Folding | | | Folding diagonally | | |
| | 10 | 100 | 1000 | 10 | 100 | 1000 | 10 | 100 | 1000 | 10 | 100 | 1000 |
| Conv-MLP | 0.0 | 5.0 | 12.5 | 10.0 | 17.5 | 22.5 | 2.5 | 7.5 | 10.0 | 0.0 | 5.0 | 15.0 |
| GCTN | 10.0 | 10.0 | 55.0 | 32.5 | 50.0 | 55.0 | 15.0 | 22.5 | 45.0 | 17.5 | 30.0 | 62.5 |
| Graph transporter | 20.0 | 30.0 | 70.0 | 45.0 | 65.0 | 75.0 | 40.0 | 60.0 | 80.0 | 45.0 | 60.0 | 65.0 |
| MLP | 0.0 | 2.5 | 10.0 | 5.0 | 17.5 | 20.0 | 0.0 | 10.0 | 12.5 | 0.0 | 7.5 | 17.5 |
| GCN | 5.0 | 7.5 | 17.5 | 7.5 | 20.0 | 35.0 | 12.5 | 15.0 | 20.0 | 0.0 | 10.0 | 15.0 |
| Local-GNN (ours) | **30.0** | **67.5** | **95.0** | **47.5** | **55.0** | **90.0** | **75.0** | **80.0** | **100.0** | **40.0** | **60.0** | **100.0** |

The average success rates (%) on unseen deformable rearranging tasks, where models are trained with 10, 100 and 1000 demonstrations per task. The best performance in each task is in bold

All above methods are provided with visual observations of the current object states and a goal visual observation as inputs, and output the SE(2) poses of robot pick and place actions at each timestep. Note that all above methods are trained in the *single-task* learning scenario.

**Success rate**

We compare above methods on their success rates of completing a common set of deformable rearrangement tasks. Specifically, we define a task as a *success* if the robot completes the task within twenty pick and place actions, and otherwise as a failure. We evaluate the success rate for each single type of deformable rearrangement tasks, with forty random *unseen* task instances, and apply all methods trained with 10, 100 and 1000 demonstrations separately.

The results are shown in Table 4. Overall, GCTN preforms better than Conv-MLP, especially for models trained with more demonstrations (e.g. the last column of each type of tasks), which demonstrates the effectiveness of the transporter architecture on deformable rearrangement. By introducing a handcrafted graph structure to provide global interaction information, Graph Transporter outperforms GCTN, which illustrates the necessity of global interactions in deformable object rearrangement.

Leveraging the dynamic graph representation and local-GNN based policy learning model, our method outperforms all baseline methods with the highest success rates on all task cases. Particularly on the tasks with more twining goal configurations (e.g. the N-shape and Square-shape scenarios of rope manipulation) or tasks with more complex deformable dynamics (e.g. all cloth manipulation tasks), our model shows more significant advantages of achieving higher success rates than the baseline models. It should also be mentioned that our proposed local-GNN model outperforms other models based on keypoints (MLP, GCN). In our local-GNN, we adopted an attention mechanism to learn the interaction relationships (whether there is an edge between every two nodes and the weight of the edge), which leads our model to outperform MLP and GCN (based on handcrafted complete graphs).

Our method also achieve orders of magnitude higher sample efficiency than baseline models. Particularly, it can be observed in Table 4 the success rates of our method trained on 10 and 100 demonstrations are much higher than those of baseline models trained on 10 and 100 demonstrations, and even higher than those of the baselines trained on 1000 demonstrations on almost all involved tasks.

These results demonstrate that compared with other architectures, our local-GNN model architecture can capture the dynamics of deformable rearrangement more efficiently and accurately, and therefore can be a more general and suitable framework for deformable rearrangement tasks. In other words, our model is more expressive than the baseline methods in modeling the deformable rearrangement dynamics. Figure 7 shows several example solutions of goal-conditioned deformable rearrangement tasks generated by our method trained on 1000 demonstrations. Note that Fig. 7

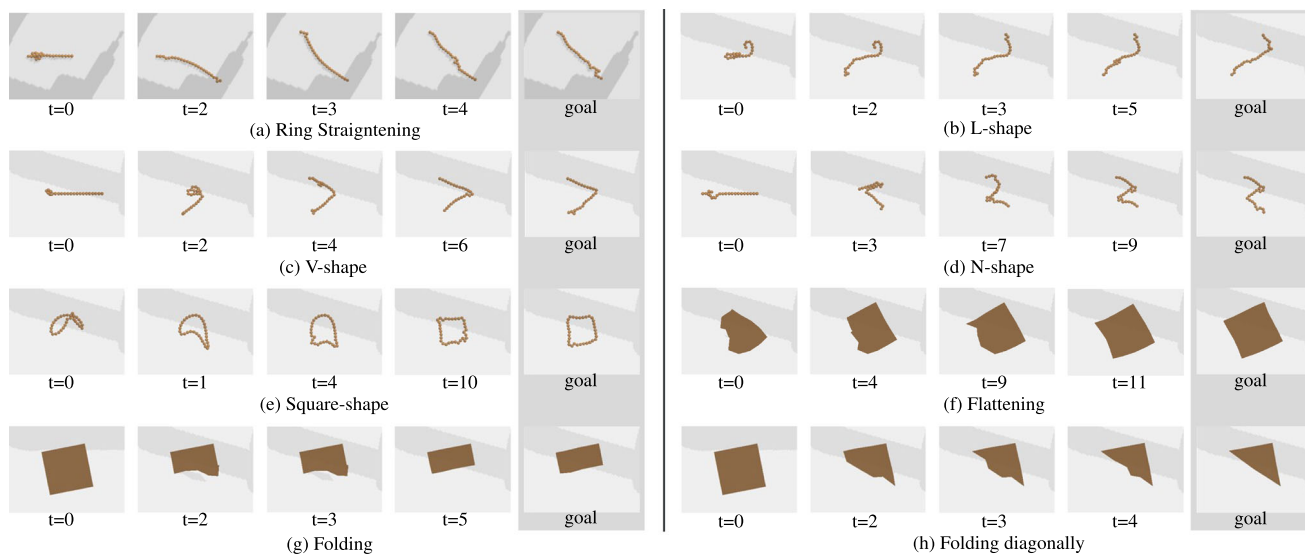

**Fig. 7** We evaluate our framework on eight types of deformable rearrangement tasks in simulation. Each example task shows four frames in the sequence. Experimental results show that our framework captures the dynamics of deformable rearrangement accurately and efficiently

shows only key action frames for the simplicity of demonstration, but all tasks are completed in twenty actions.

**Model capacity**

Besides having a superior expressiveness on the deformable rearrangement problem, our model is much lighter and simpler than the baseline methods. For a detailed comparison, we calculate the FLOPs (Floating Point Operations), model parameters and inference times of our model and GCTN. As explained in "Learning for deformable object rearrangement", our method consists of a keypoint detector and a local-GNN running in a close-loop manner. For a fair comparison, the inference time of our framework is therefore calculated as the total time of detecting keypoints and inferring manipulation actions from the keypoints. We run the analysis with an NVIDIA Tesla T4 GPU on an Intel Xeon Platinum 8255C CPU. The results are shown in Table 5. Overall, our framework is dramatically (more than two orders of magnitude) lighter in terms of model FLOPs and parameters than GCTN, which also leads to much less (more than two times) inference time of our model.

The combined strength of expressiveness, efficiency and simplicity of our model mainly comes from that compared with the convolutional visual features centered in many previous methods [3,16], the graph features highlighted in our model excel at handling the sparse information of deformable dynamics in rearrangement and can capture the characteristics of deformable rearrangement in a more accurate and efficient manner. In addition, the numerical calculations on two keypoint sets (while each set is essentially a much smaller subset of the whole pixel set of a visual observation) in our model are much less than those of convolution operations on two whole visual observations, which further benefits the efficiency of our proposed framework.

**Sim-to-real experiments**

As described previously, the processing of visual observations (keypoint detection) in our framework is naturally separated from the subsequent planning of manipulation actions (local-GNN update and action generation). Such a hierarchy feature enables the transfer of our framework from simulation to reality much more simple and robust. Specifically, our framework can learn to rearrange deformable objects from a large quantity of demonstrations in simulation at the first stage. Since the planning module with local-GNN takes only keypoints as inputs, our framework can transfer the learned skills from simulation to reality by only fine-tuning the keypoint detector, i.e. to accommodate the sim-to-real gap in the visual observations. In addition, since our planning module is a GNN-based model architecture, the length of each input sequence (i.e. the number of keypoints in each dynamic graph) is adjustable, without the need of modifying the model architecture, which further simplifies the sim-to-real of our framework.

We evaluate the sim-to-real performance of our framework with a UR5 robot manipulator mounted with a suction cup (Fig. 8). The deformable object (either a rope or a cloth) is placed on the platform in front of the robot. Note that we have used a cloth with a plaid pattern, which is used for the ease of fine-tuning the keypoint detector in real images. The plaid pattern can help us label keypoints in the image manually, which can increase the efficiency of keypoint detector

**Table 5** Analysis on the model capacity of different models

| Model | FLOPs (M) | Parameters (K) | Inference time (s) |
| --- | --- | --- | --- |
| GCTN | 326,249 | 39,337 | 0.350 |
| Our pipeline[a] | **1844** | **575** | **0.145** |
| Keypoint detector | 910 | 75 | 0.060 |
| Local-GNN | 24 | 500 | 0.025 |

The bold texts highlight our results and show that our method outperforms the compared method

[a]Our pipeline consists of a keypoint detector and a local-GNN

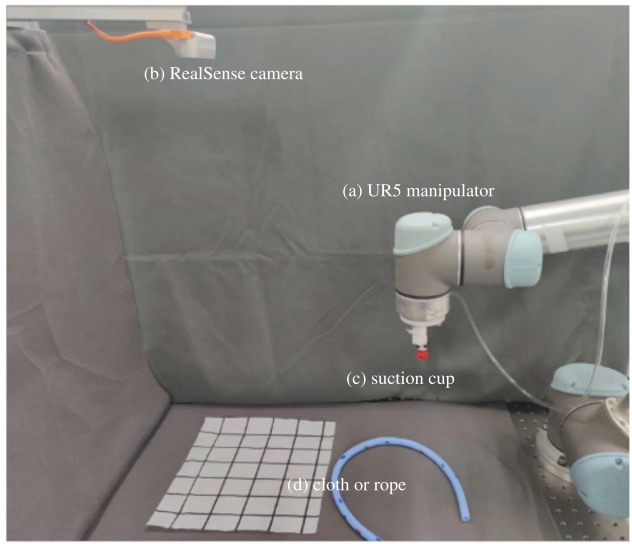

**Fig. 8** Our real experimental setup includes **a** a UR5 robot, **b** a Realsense camera, **c** a suction cup and **d** a deformable object (rope or cloth)

fine-tuning. We use a Realsense RGB-D camera fixed on the top of the platform and just above the deformable object to collect visual observations. We apply a model instance of our framework initially trained on 1000 demonstrations in simulation. We then collect 500 real visual observations of the rope and the cloth each, and use them to fine-tune the sim-trained keypoint detector to reduce the sim-to-real gap, e.g. the errors caused by image style difference. We set an empirical keypoint number for the involved rope and cloth rearranging tasks to be 5 and 8 respectively, which however can be determined differently, e.g. proportional to the complexity of the task or object dynamics.

We apply the fine-tuned model on a number of twenty random instances for both rope and cloth rearrangement tasks. The obtained average success rates are 100% and 95% for the rope and cloth rearrangement respectively, which are almost comparable to the model performance in simulation (Table 4). Figures 9, 10, 11, 12, 13 and 14 show six example deformable rearrangement tasks in our real experiments. Each task is provided with an initial and a goal visual observation (the leftmost sub-figure of each task). The final visual state of the deformable object is illustrated and compared with the provided goal state (the rightmost sub-figure of each task). Clearly, our model completes all involved real tasks with a final object state very close to the provided goal. At each timestep, we show the selected optimal pick and place actions, and their corresponding probability distributions over the object keypoints respectively. A full video recording of these experiments can be found in the supplementary video.

**Multi-task policy learning**

Most previous learning frameworks for deformable rearrangement are typically performed in the *single-task* learning scenario [3,26], i.e. one model instance is supposed to be trained separately to learn the specific manipulation skill for each single type of deformable objects. Such a task isolation in face results from their limited efficiency and generalization in handling deformable rearrangement tasks. In consideration of the superior expressiveness and simplicity of our method, we also train our model in the *multi-task* learning scenario, i.e. we train one single model with demonstrations from all different types of deformable rearrangement tasks together in our dataset. We hope that our model can learn the rearrangement skills generalized over different types of deformable objects embedded in the *multi-task* demonstrations.

We train our *multi-task* model and then evaluate the model performance on each type of deformable rearrangement tasks separately. The results are summarized in Table 6. Comparing with the results in Table 4, the results for *single-task* learning, the *multi-task* model performs comparably to the models trained in the *single-task* learning scenario. That is, in addition to *single-task* deformable rearrangement skills, our model can learn directly *multi-task* skills of deformable rearrangement from demonstrations. It indicates that on the one hand, the rearrangement skills of different types of deformable tasks do have similarities that can be effectively captured by our model, so that our *multi-task* model can still perform well beyond the *single-task* learning paradigm. On the other hand, our graph representation strategy is sufficiently expressive and suitable for modeling more general deformable rearrangement dynamics, and thus learning deformable rearrangement skills with the graph representation has a strong generalization capability.

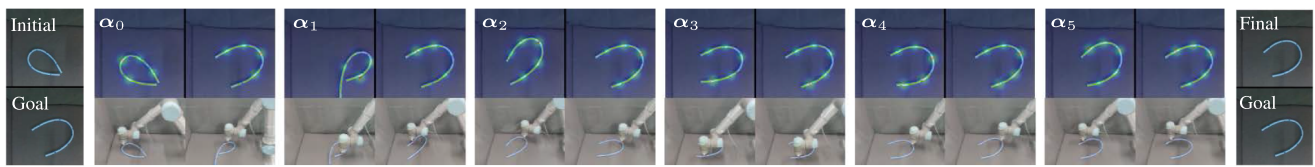

**Fig. 9** The robot rearranges a rope to a U-shape with six pick and place actions

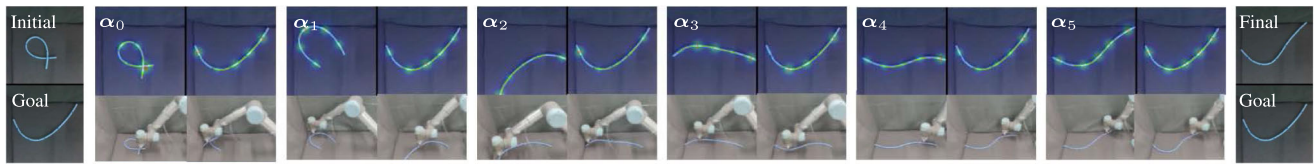

**Fig. 10** The robot rearranges a rope to a L-shape with six pick and place actions

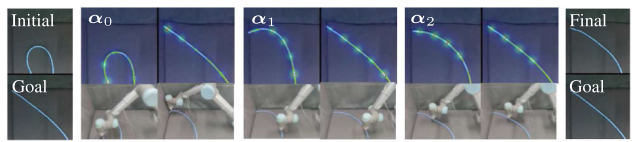

**Fig. 11** The robot straightens a bent rope with three pick and place actions

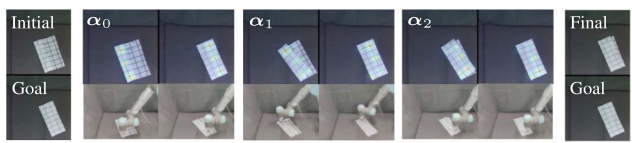

**Fig. 12** The robot folds the cloth in half with three pick and place actions

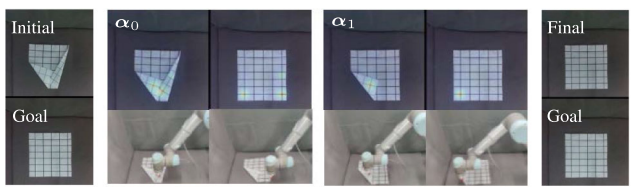

**Fig. 13** The robot flattens the cloth with three pick and place actions

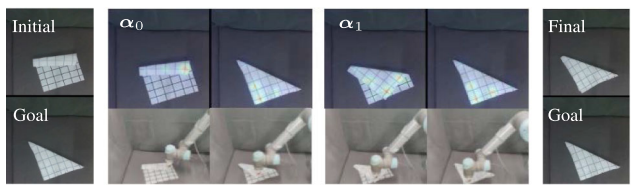

**Fig. 14** The robot folds the cloth diagonally with three pick and place actions

**Table 6** The average success rates (%) of multi-task learning in simulation by our proposed framework

| Straightening | V-shape | L-shape | N-shape |
| --- | --- | --- | --- |
| 100.0 | 100.0 | 100.0 | 85.0 |
| Square-shape | Flattening | Folding | Folding diagonally |
| 95.0 | 90.0 | 100.0 | 100.0 |

## Conclusion

We have proposed local-GNN, a light and efficient learning framework for vision-based goal-conditioned deformable object rearrangement tasks. Different from many previous studies which rely mainly on the convectional features from visual observations [3,17,18], our method leverages keypoints and dynamic graphs to model the deformable rearrangement dynamics. Extensive experiments have been conducted to demonstrate the performance of proposed dynamic graph structures in deformable object representation, and prove that our local-GNN can be more general and suitable for learning goal-conditioned deformable rearrangement policies. One limitation of our model is that our model learns manipulation policy directly from keypoints. However, when the deformable object has self-occlusion, it would be hard to detect accurate keypoints from RGB images. In future work, we aim to investigate detecting keypoints from point clouds rather than RGB images, which can potentially broaden the application scenarios of the proposed method.

**Data Availability** The code will be open-source on https://github.com/dengyh16code/deformable-gnn.git in a week.

### Declarations